%% file: anonymous-submission-latex-2025.tex
\documentclass[letterpaper]{article} 
\usepackage{aaai25}  
\usepackage{times}  
\usepackage{helvet}  
\usepackage{courier}  
\usepackage[hyphens]{url}  
\usepackage{graphicx} 
\usepackage{amsmath} 
\usepackage{booktabs}
\usepackage{siunitx} 
\usepackage{makecell}
\usepackage{caption}
\usepackage{natbib}  
\usepackage{caption} 
\usepackage{algorithm}
\usepackage{algorithmic}
\usepackage{enumitem}
\nocopyright
\usepackage{newfloat}
\usepackage{listings}
\DeclareCaptionStyle{ruled}{labelfont=normalfont,labelsep=colon,strut=off} 
\floatstyle{ruled}
\newfloat{listing}{tb}{lst}{}
\floatname{listing}{Listing}
\title{
TN-AutoRCA: Benchmark Construction and Agentic Framework for Self-Improving Alarm-Based Root Cause Analysis in Telecommunication Networks
}
\author {
    Keyu Wu$^{\mathrm{1}}$,
    Qianjin Yu$^{\mathrm{1}}$,
    Manlin Mei$^{\mathrm{1}}$,
    Ruiting Liu$^{\mathrm{1}}$,
    Jun Wang$^{\mathrm{2}}$,
    Kailai Zhang$^{\mathrm{2}}$,
    Yelun Bao$^{\mathrm{2}}$, \\
    Wenbo Wan$^{\mathrm{1}}$,
    Rui Ye$^{\mathrm{1}}$,
    Bin He$^{\mathrm{1}}$,
    Junfeng Liao$^{\mathrm{1}}$, 
    Yongsheng Du$^{\mathrm{1}}$,
    Zhiguo Yang$^{\mathrm{1}}$,
    Kunlin Liu$^{\mathrm{1}}$,
    Lingjun Huang$^{\mathrm{1}}$,
    Zhiwei Song$^{\mathrm{1}}$, 
    YanQin Gao$^{\mathrm{1,\ddagger}}$,
    Fang Tan$^{\mathrm{1,\ddagger}}$
    Jin Yang$^{\mathrm{2,\ddagger}}$,
    Ninglun Gu$^{\mathrm{2,\ddagger}}$
}
\affiliations {
    \textsuperscript{\rm 1}\textbf{Zhongxing Telecom Equipment(ZTE), China}\\
    \textsuperscript{\rm 2}\textbf{China Mobile Communications Group Co Ltd}\\
    AIM@zte.com.cn, \{wangjunwl,zhangkailai,baoyelun,yangjin,guninglun\}@chinamobile.com
}
\usepackage{bibentry}

\begin{document}
\maketitle
%
\input{abstract}
\input{introduction}

\input{related_work}
\input{method}

\input{experiments}
\input{conclusion}



\bigskip

\bibliography{aaai25}

\end{document}

%% file: abstract.tex
\begin{abstract}
Root Cause Analysis (RCA) in telecommunication networks is a critical task, yet it presents a formidable challenge for Artificial Intelligence (AI) due to its complex, graph-based reasoning requirements and the scarcity of realistic benchmarks. To catalyze research in this domain, We herein present \textbf{TN-RCA530}, the inaugural real-world, publicly accessible benchmark for root cause analysis (RCA) of telecommunication network alarms, comprising 530 fault scenarios constructed from expert-validated Knowledge Graphs(KGs). Our evaluation reveals that even state-of-the-art Large Language Models (LLMs) perform poorly on this task, with the best models achieving an F1-score below 70\%, highlighting its significant difficulty.To address this challenge, we then propose \textbf{Auto-RCA}, a novel agentic system that automates the iterative refinement of a code-based solution. The core innovation of Auto-RCA lies beyond simple self-correction; it employs an iterative "evaluate-analyze-repair" loop that systematically identifies common patterns across all failure cases to generate contrastive feedback. This feedback guides the LLM to fix systemic logical flaws rather than isolated errors. Experiments show that this agentic framework dramatically boosts problem-solving performance, elevating the final solution's F1-score on TN-RCA530 from a baseline of 58.99\% (achieved by Gemini-2.5-Pro directly) to 91.79\%. This work not only contributes a crucial benchmark to the community but also demonstrates that autonomous, self-optimizing agentic architectures are a powerful paradigm for solving complex, domain-specific reasoning problems.
\end{abstract}

%% file: introduction.tex
\section{introduction}

\begin{figure}[t]
\centering
\includegraphics[width=\columnwidth]{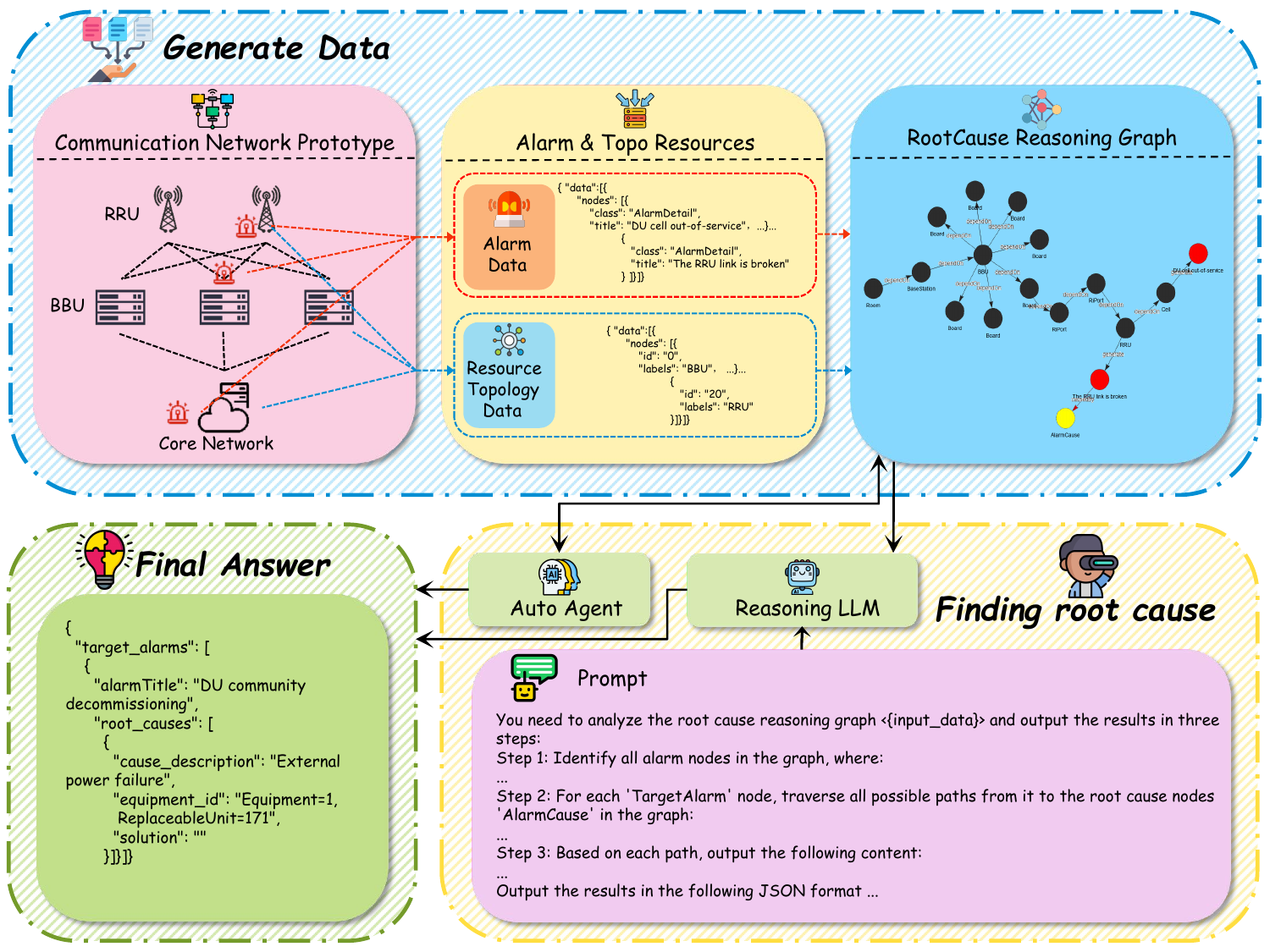} 
\caption{The framework for LLM-based root cause analysis in telecommunication base stations. The process begins by modeling real-world base station equipment (e.g., RRU, BBU) and their connections using a knowledge graph. This graph, along with real-time alarm data and expert-defined prompts, is fed into a Large Language Model (LLM). The LLM then leverages its reasoning capabilities to perform root cause analysis, ultimately generating a structured output that includes a description of the root cause, the specific equipment involved, and a proposed solution.}
\label{fig1}
\end{figure}

Alarm-Based Root Cause Analysis (RCA) is a cornerstone of reliability engineering in modern telecommunication networks, essential for ensuring network resilience and minimizing service downtime. 
However, traditional approaches to alarm root cause analysis rely critically on domain experts to perform detailed analysis of intricate alarm log data for accurate root cause identification. This reliance introduces significant subjectivity tied to individual experience, hindering the establishment of unified data management protocols and systematic modeling frameworks.
Recent advances in large language models (LLMs) have yielded remarkable progress in reasoning capabilities, with state-of-the-art models achieving human-level or superior performance on mathematical and coding benchmarks~\cite{deepseekai2025deepseekr1incentivizingreasoningcapability,yang2025qwen3technicalreport, qwq32b, kimiteam2025kimik15scalingreinforcement}. This breakthrough renders LLMs viable for tackling complex real-world problem-solving.
Drawing inspiration from the fundamental topological interdependencies of communication network infrastructure, we introduce an innovative methodology for unified data modeling and generation. Our solution is depicted in Figure ~\ref{fig1}, integrates alarm RCA-critical raw alerts and topological resources via a knowledge graph, subsequently synthesizing them into a root cause inference graph.

Unlike common reasoning tasks, telecommunication network alarm-based RCA is fundamentally a multi-label classification problem over complex, graph-structured data.
It demands sophisticated reasoning based on deep domain knowledge to pinpoint the origin of cascading failures amidst a flood of noisy, interdependent alarms. 
A promising approach to automate this process involves integrating the structured knowledge representation of Knowledge Graphs (KGs) with the reasoning capabilities of Large Language Models (LLMs). 
This paradigm, illustrated in Figure~\ref{fig1}, models the network's physical equipment and their connections as a KG, which, along with real-time alarm data, is fed to an LLM to deduce the root cause.

Despite the conceptual appeal of this framework, its practical efficacy has been largely unverified. The advancement of AI-driven RCA solutions has been hampered by two significant gaps. First, there is a salient lack of comprehensive, real-world benchmarks. Existing benchmarks like SWE-bench~\cite{arora2024masaimodulararchitecturesoftwareengineering}  or OpenRCA~\cite{xu2025openrca}  focus on general software engineering and do not capture the unique challenges of the telecom domain, such as "alarm storms" and the intricate dependencies between physical and logical network layers introduced by 5G and NFV. Second, and as a consequence, the true capability of LLMs on this task has been an open question.

To quantify this challenge and establish a clear baseline, we evaluated a suite of prominent LLMs on a new, real-world telecom benchmark we developed. The results are sobering. This direct application of LLMs—a measure of their raw, intrinsic reasoning ability on the task—reveals a significant performance ceiling. Even a state-of-the-art model achieves a top F1-score of only 58.99\%. This finding underscores that merely scaling up general-purpose models is insufficient for mastering this complex domain and validates the difficulty of the benchmark itself.

To bridge these gaps, this paper presents a dual contribution that first rigorously defines the problem space and then offers a robust method for solving it. We introduce \textbf{TN-RCA530}, the first large-scale, public benchmark specifically designed for telecom RCA, providing a standardized "proving ground" to measure and drive progress. Subsequently, we present \textbf{Auto-RCA}, a novel agentic framework that demonstrates a path to mastery on this benchmark. Rather than enhancing the LLM's inherent abilities, Auto-RCA leverages the model as a reasoning engine within a larger system. This system iteratively refines a code-based solution by systematically analyzing patterns across all failures to generate targeted, contrastive feedback for the LLM.

Our work makes the following key contributions:
\begin{itemize}[leftmargin=*]
    \item \textbf{A Novel Telecom RCA Benchmark (TN-RCA530):} We introduce the first large-scale, real-world benchmark for telecom RCA, featuring 530 scenarios grounded in expert-verified knowledge graphs, a verifiable construction methodology, and an automated complexity grading system.
    
    \item \textbf{An Autonomous Agent with a Contrastive Feedback Loop (Auto-RCA):} We design and implement Auto-RCA, a modular agentic system that iteratively improves a code-based solution. Its core innovation is a novel "evaluate-analyze-repair" loop that generates powerful, structured contrastive feedback by systematically analyzing aggregate failure patterns. This guides the LLM to perform targeted, hypothesis-driven code repair, a significant advance over simple self-reflection.

     \item \textbf{State-of-the-Art Problem-Solving Capability:} Through extensive experiments, we show that the Auto-RCA framework dramatically improves the ability to solve the tasks in our benchmark, elevating the final solution's F1-score on TN-RCA530 to 91.79\%. This establishes a new state-of-the-art performance on this specific problem and charts a viable path toward automating complex, domain-specific reasoning.
\end{itemize}

%% file: related_work.tex
\section{Related Work}

\subsection{Knowledge Graph Enhanced LLMs}
Integrating Knowledge Graphs (KGs) with Large Language Models (LLMs) is a key strategy to improve factual grounding and mitigate hallucinations~\cite{Ibrahim2024survey, make7020038}. Current research focuses on frameworks that align LLM reasoning with KG paths~\cite{shen2025reasonalignrespondaligningllmreasoning}, use LLMs as policy models to navigate KGs~\cite{shenetal}, or leverage KGs to rerank generated candidates~\cite{electronics13245037}. These approaches highlight a trend towards structured neuro-symbolic reasoning, which is essential for complex, fact-based tasks like RCA~\cite{amayuelas2025groundingllmreasoningknowledge}.

\subsection{Root Cause Analysis (RCA)}
In software engineering, AI-driven RCA increasingly replaces manual methods by leveraging multi-modal data from logs, metrics, and traces~\cite{pham2025rcaevalbenchmarkrootcause, elafrca, xu2025openrca}. While general benchmarks like Defects4J~\cite{Just2014Defects4J} and RCAEval~\cite{pham2025rcaevalbenchmarkrootcause} have been vital, they do not address the unique challenges of telecommunication networks. Telecom RCA is distinct due to massive "alarm floods"~\cite{zhang2021influencebasedapproachrootcause}, complex topological dependencies, and stringent real-time constraints~\cite{ararci, rttnm}. Despite advances in applying AI/ML to this domain~\cite{dfnn, en17225758, pr12112312, Liang2020ABN, acafiicn}, the field has lacked a large-scale, public benchmark, a gap our \textit{TN-RCA530} benchmark is designed to fill.

\subsection{LLM-based Agents for Automated Program Repair}
The paradigm of using LLMs as autonomous agents for Automated Program Repair (APR) has shown significant promise, with benchmarks like SWE-bench becoming a standard for evaluation~\cite{campos2025empiricalevaluationgeneralizableautomated, ma2025thinkinglongerlargerenhancing}. A prominent architectural trend is the use of modular agentic systems, where specialized sub-agents collaborate to solve complex tasks, a philosophy adopted by frameworks like MASAI~\cite{arora2024masaimodulararchitecturesoftwareengineering} and our Auto-RCA.

A critical aspect of agentic APR is the mechanism for self-correction. Early approaches like Self-Refine~\cite{pan-etal-2024-automatically}, which rely on an LLM's intrinsic ability to fix its own output, have proven ineffective for complex reasoning tasks due to "cognitive fixedness"~\cite{kamoi-etal-2024-llms}. The consensus is that effective improvement requires reliable external feedback. Building on this, methods like ContrastRepair provide contrastive feedback using pairs of failing and passing tests to guide the LLM~\cite{kong2024contrastrepairenhancingconversationbasedautomated}. Our Auto-RCA framework advances this concept by orchestrating an LLM to move beyond such instance-level fixes. The Auto-RCA system analyzes failure patterns across the entire benchmark to generate more holistic, strategic feedback. This feedback then guides the LLM in repairing the solution code to address systemic flaws, demonstrating how a structured agentic process can achieve results that direct LLM application cannot.

\subsection{LLMs for Telecommunication Networks}
LLMs are increasingly being explored to transform telecom operations, but general-purpose models often lack the required domain-specific knowledge~\cite{maatouk2025telellmsseriesspecializedlarge, ethiraj2025efficienttelecomspecificllm, otellm}. This has led to the development of specialized "Telco LLMs" and tool-assisted agents. For instance, frameworks like TAMO use agents with multi-modal data for fine-grained RCA in AIOps~\cite{wang2025tamofinegrainedrootcauseanalysis}. This highlights that effective and reliable RCA in dynamic environments requires grounding LLM reasoning with external tools or structured knowledge. Our work contributes to this direction by providing both a rigorous benchmark (TN-RCA530) to evaluate performance on this task, and an advanced agentic framework (Auto-RCA) that demonstrates how to effectively apply an LLM to achieve state-of-the-art problem-solving performance.

%% file: method.tex
\section{Methodology}
Our methodology addresses the challenge of telecommunication root cause analysis through a two-part contribution. First, to rigorously define and measure the problem, we introduce \textbf{TN-RCA530}, a novel and challenging benchmark grounded in real-world data. Second, to solve the problem, we present \textbf{Auto-RCA}, an agentic framework designed to iteratively learn and master the complexities of the benchmark. This section details the construction principles of \textbf{TN-RCA530}, the architecture of the \textbf{Auto-RCA} system, and the unified evaluation framework used to assess performance.

\subsection{Benchmark Construction: The TN-RCA530 Dataset}

The primary contribution of our work is TN-RCA530, a benchmark comprising 530 distinct tele-network failure scenarios. Its construction was governed by four core principles: Veracity, Comprehensiveness, Verifiability, and Complexity Discriminability. The input for each scenario is a knowledge graph that models the physical equipment, their topological connections, and associated alarm data. The LLM's task is to leverage its domain knowledge and reasoning abilities to analyze this graph and identify the specific root cause node(s) from a set of potential candidates.

To achieve these principles, we adopted a result-oriented construction process, which we detail below.

\textbf{Veracity}: To ensure the benchmark reflects real-world challenges, all data is sourced directly from operational telecommunication base stations. This includes the network topology, which describes the physical and logical connections between equipment (e.g., Baseband Units (BBUs), Remote Radio Units (RRUs)), and the alarm data, which are real signals generated during actual fault events.

\textbf{Comprehensiveness}: A brute-force approach of collecting all possible fault scenarios is inefficient and prone to gaps. Instead, we employ a result-oriented methodology. We begin by identifying a comprehensive set of known root causes based on industry standards and expert knowledge. We then work backward from these root causes, tracing their potential alarm manifestations through the network topology. This ensures a broad and diverse coverage of root cause types, capturing both common and long-tail fault scenarios more effectively.

\textbf{Verifiability} : The result-oriented approach provides an inherent ground truth for each scenario. Since every case is constructed by tracing from a known, expert-verified root cause to its resulting alarms, the correct answer (the root cause) is embedded in the data by design. This provides a reliable and unambiguous basis for evaluating model performance.

\begin{figure}[htbp]
\centering
\includegraphics[width=\columnwidth]{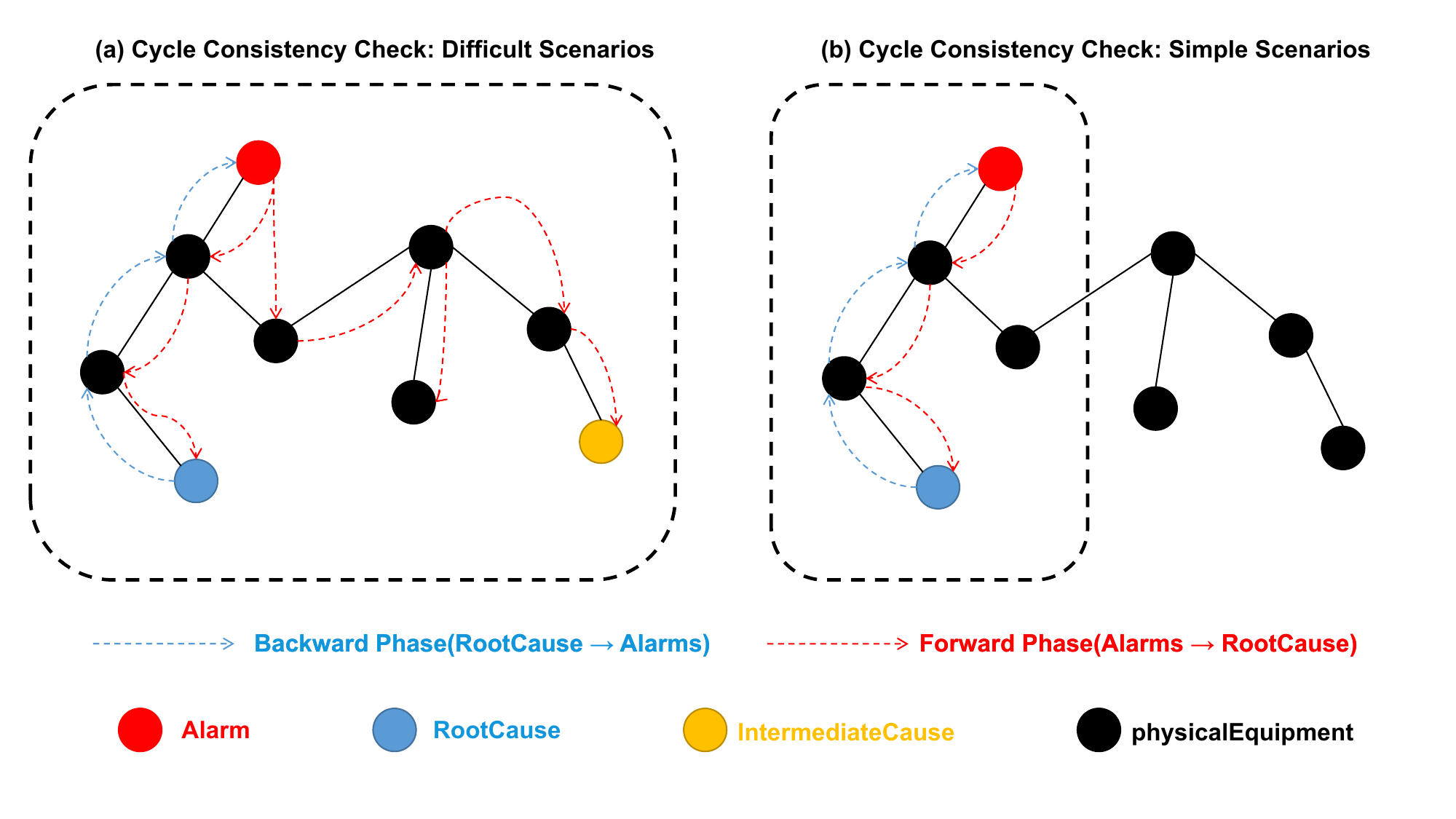} 
\caption{Illustration of the cycle consistency check for automated difficulty grading. (a) A \textbf{Difficult Scenario}, where the forward reasoning phase (red arrows, Alarms $\to$ RootCause) results in a one-to-many mapping ($|m|>1$). The initial Alarm can be traced to both the true \textit{RootCause} and other plausible causes (e.g., \textit{IntermediateCause}), creating ambiguity. (b) A \textbf{Simple Scenario}, where the forward mapping is one-to-one ($|m|=1$). The causal path from the Alarm to the RootCause is unambiguous, and the backward phase (blue arrows, RootCause $\to$ Alarms) confirms this unique, consistent cycle.}
\label{fig:cycle_consistency}
\label{fig2}
\end{figure}

\begin{figure*}[htbp]
\centering
\includegraphics[width=\textwidth]{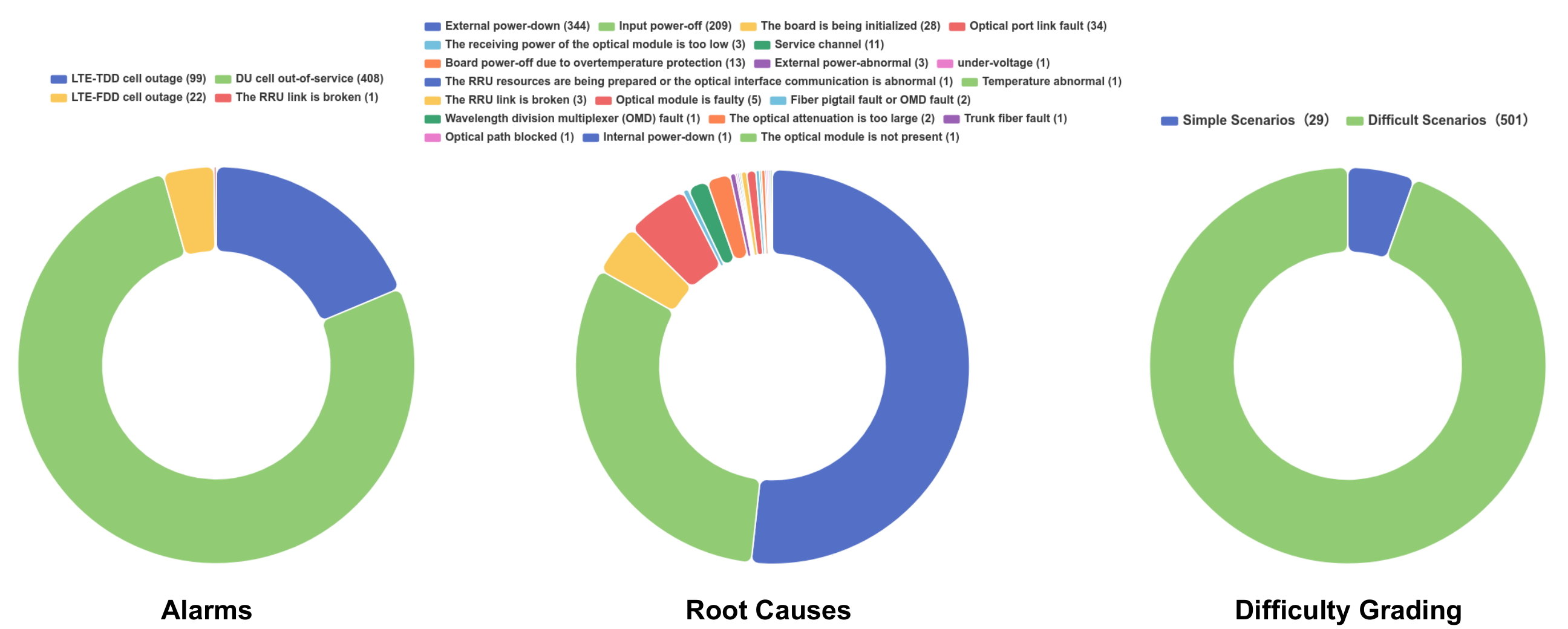} 
\caption{Distribution and Complexity Stratification of Root Causes in TN-RCA530. The chart displays the frequency of major root cause categories, illustrating the benchmark's long-tail distribution. Each bar is segmented to show the absolute number of 'Simple' versus 'Difficult' scenarios for that cause. This visualization reveals two key properties: (1) the dominance of a few common faults, and (2) the overwhelming prevalence of 'Difficult' scenarios (94.5\% of the total) across all categories, highlighting the benchmark's high level of challenge.}
\label{fig:root_cause_distribution}
\end{figure*}

\textbf{Complexity Discriminability}: A key innovation of our methodology is the ability to objectively and automatically stratify the difficulty of scenarios. This is achieved through a cycle consistency check between a backward and a forward reasoning process, grounded in the graph structure.

\begin{itemize}
    \item \textbf{Backward Phase (Root Cause → Alarms):}This phase establishes the ground truth. For a given standard root cause (e.g., "External Power Failure"), we utilize the resource topology graph and expert-defined rules to determine the complete chain of alarms that this cause would trigger. This creates a definitive mapping:

    \begin{equation}
    \begin{split}
        M_{backward}:RootCause \to \qquad \qquad \qquad \qquad \\
        \left \{ Alarm_1,Alarm_2,\dots ,Alarm_n \right \}
    \end{split}
    \end{equation}

    \item \textbf{Forward Phase (Alarms → Root Cause):} This phase simulates the LLM's task. Starting from the set of alarms identified in the backward phase, we construct the input knowledge graph. We then apply graph-based reasoning to generate a potential inference graph that links the alarms back to one or more possible root causes. This establishes a forward mapping:
    \begin{equation}
    \begin{split}
        M_{forward}:{Alarms} \to \qquad \qquad \qquad \qquad \qquad \qquad \\
        \left \{ InferredRootCause_1,\dots ,InferredRootCause_m \right \}
    \end{split}
    \end{equation}
 The uniqueness of causal paths in the graph provides a theoretical basis for this process.
    
    \item \textbf{Cycle Consistency Check \& Difficulty Grading:}  We assess consistency by comparing the output of the forward phase with the input of the backward phase.
    \begin{itemize}
    \item \textbf{Simple Scenarios:}A scenario is classified as "simple" if there is a one-to-one mapping ($\left | m \right | =1$) and the inferred root cause perfectly matches the original root cause. The causal path from alarms to the root cause is unambiguous.

    \item \textbf{Difficult Scenarios:}A scenario is classified as "difficult" if there is a one-to-many mapping ($\left | m \right | >1$), where the alarms could plausibly point to multiple root causes, including the ground truth and several distractors. The difficulty level is proportional to the number of plausible alternative root causes, as this requires the model to perform more nuanced reasoning to eliminate incorrect hypotheses.
    
    \end{itemize}

\end{itemize}
This automated process removes subjective manual labeling of difficulty and provides a principled mechanism for analyzing model performance across a spectrum of complexity.

\subsection{Data Representation and Statistics}

Each scenario in the \textbf{TN-RCA530} benchmark is encapsulated in a pair of JSON files: \textit{input.json}, which contains the knowledge graph representation of the fault scenario for the LLM to analyze, and \textit{label.json}, which provides the ground-truth answer.

\textbf{Knowledge Graph Structure: }The \textit{input.json} file models a fault scenario as a directed graph $G=(V,E)$, where $V$ is a set of nodes and $E$ is a set of edges representing their relationships. The nodes $V$ are primarily categorized into three types, essential for the RCA task:

\begin{itemize}
    \item \textbf{Resource Topology Nodes:} These nodes represent the physical and logical components of a telecommunication base station, such as the Base Station (\textit{BaseStation}), Baseband Unit (\textit{BBU}), specific functional boards (\textit{Board}, e.g., VBPe5a), Remote Radio Unit (\textit{RRU}), and ports (\textit{RiPort}). Each node contains properties like its logical distinguished name (\textit{ldn}) and serial numbers, forming the structural backbone of the network.

    \item \textbf{Alarm Nodes (\textit{AlarmDetail}):} These nodes represent the symptom data. They are linked to specific resource nodes, indicating which piece of equipment reported an alarm. Each alarm node contains rich metadata, including the \textit{title} (e.g., "LTE Cell Out of Service"), \textit{code}, \textit{severity}, and the exact \textit{reportAlarmTime}.

    \item \textbf{Root Cause Nodes (\textit{AlarmCause})}: These nodes represent the potential underlying causes of the alarms. In the \textit{input.json}, these are candidate causes linked to resource nodes. The LLM's goal is to identify which of these candidates is the true root cause for a given \textit{TargetAlarm}. The ground-truth root cause is specified exclusively in the \textit{label.json} file for verification.

\end{itemize}

The edges $E$ in the graph define the relationships between these nodes. They include \textit{dependOn} edges to describe the physical or logical hierarchy of the equipment, \textit{generate} edges linking alarms to the equipment that produced them, and \textit{causedBy} edges that represent causal hypotheses between different alarms or between a root cause and an alarm. This structured, graph-based representation enables an LLM to perform complex multi-hop reasoning by traversing the topology and causal links to deduce the primary failure source from a cascade of symptoms.

\textbf{Benchmark Statistics:} A statistical analysis of TN-RCA530 reveals two core characteristics: a realistic root cause distribution and a challenging design dominated by "difficult" scenarios.

First, the distribution of root causes in the benchmark exhibits a typical "long-tail" pattern (as shown in Figure \ref{fig:root_cause_distribution}). A few common faults (e.g., "External power-down") account for the majority of cases, while a large number of diverse, infrequent faults make up the "tail." This distribution aligns closely with real-world operational data, ensuring the benchmark's practical relevance.

Second, and a key innovation of our work, we applied an automated "cycle consistency check" methodology to stratify all 530 scenarios by difficulty. The result is a benchmark composed of 29 "Simple" scenarios and 501 "Difficult" scenarios. This means 94.5\% of the cases are classified as "Difficult." This design establishes TN-RCA530 as a highly challenging benchmark capable of effectively differentiating the upper limits of various models' reasoning abilities, especially when handling complex and ambiguous problems that remain unsolved.

\subsection{The Auto-RCA Agentic Framework}
Having established the TN-RCA530 benchmark to quantify the challenges of telecom RCA, we now introduce \textbf{Auto-RCA}, the agentic framework we developed to master this task. Auto-RCA is not designed to enhance the LLM's intrinsic reasoning but to orchestrate its capabilities within a structured, iterative process that refines a programmatic, code-based solution by learning from its failures on the benchmark.

\textbf{Core Philosophy:} The central tenet of Auto-RCA is to treat failure not as an isolated incident but as a systemic learning opportunity. Instead of fixing bugs one by one, the framework analyzes patterns across all failure cases (Bad Cases) to identify the primary logical deficiencies in the current solution code. This enables the system to guide the LLM toward systemic, hypothesis-driven repairs rather than superficial, instance-level patches.

\begin{figure}[htbp]
\centering
\includegraphics[width=\columnwidth]{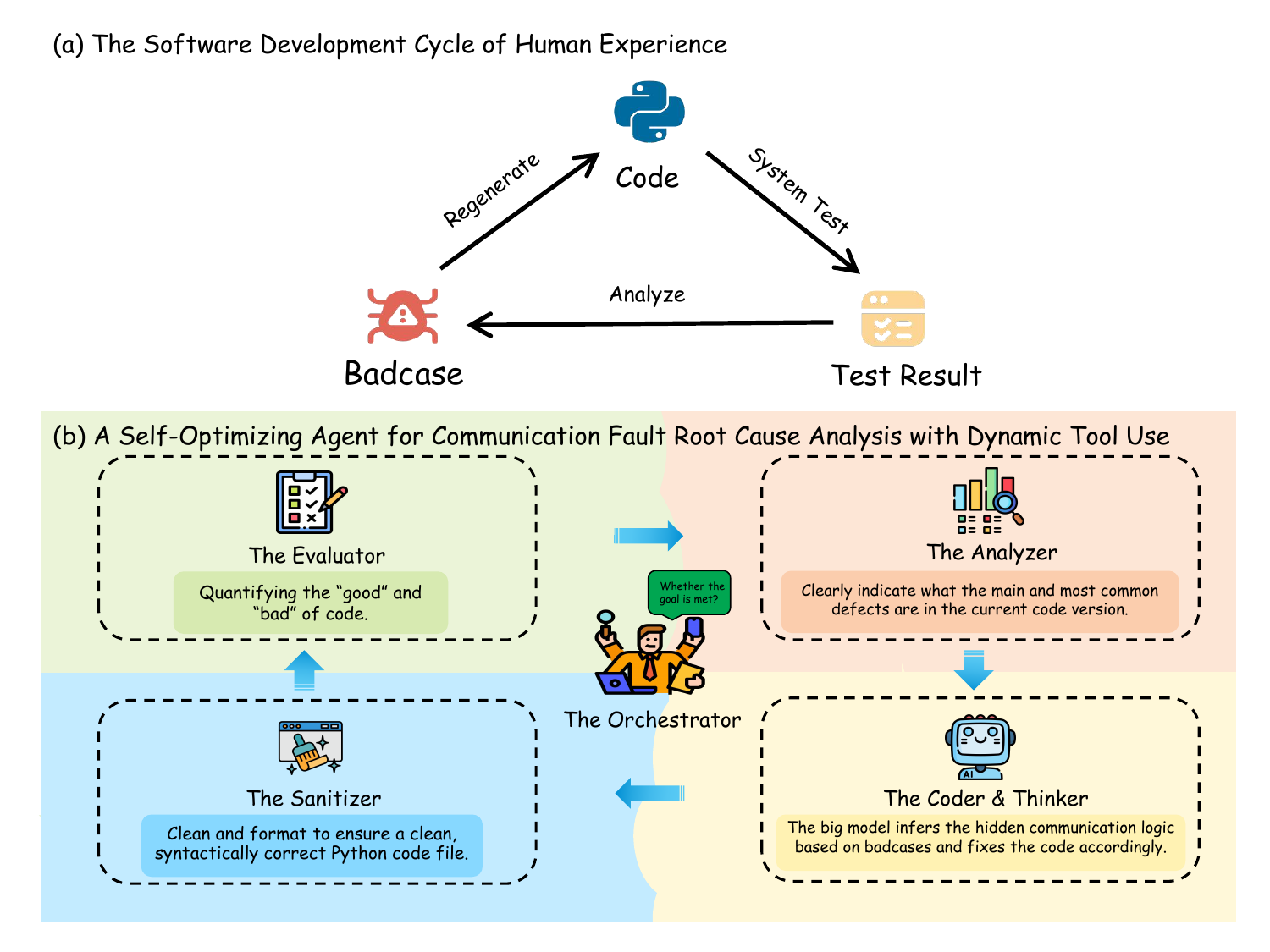} 
\caption{The architecture of the Auto-RCA agentic framework. The system operates in a continuous loop where the Orchestrator manages the workflow between modules dedicated to evaluating code performance, analyzing failures, generating targeted prompts for an LLM, and sanitizing the resulting code for the next iteration.}
\label{fig:auto-rca}
\end{figure}

\textbf{System Architecture:} As illustrated in Figure~\ref{fig:auto-rca}, Auto-RCA employs a modular architecture composed of five synergistic modules that emulate an expert software development team. This structure separates concerns and ensures a robust, repeatable workflow.
\begin{itemize}
\item \textbf{The Orchestrator (Project Manager):} Manages the end-to-end "evaluate-analyze-generate-validate" lifecycle. It controls the workflow and makes the final decision to accept or reject a proposed code modification based on empirical performance changes on the benchmark.
\item \textbf{The Evaluator (Test Engineer):} Quantifies the performance of a given code solution. It runs the code against the entire TN-RCA530 benchmark, calculates a precise F1-Score, and compiles all incorrect outputs into a set of Bad Cases for analysis.
\item \textbf{The Bad Case Analyzer (Senior Analyst):} This is the intelligent core of the framework. It receives all Bad Cases from the Evaluator and performs a systematic analysis to identify and categorize the most prevalent failure mode (e.g., "missing root cause" vs. "extra root cause"). This analysis is the foundation for the system's contrastive feedback.
\item \textbf{The LLM Agent (Coder \& Thinker):} This module acts as the interface to the Large Language Model. It translates the structured output from the Bad Case Analyzer into a high-quality, targeted prompt. This prompt combines the failure analysis report, representative Bad Cases, and pre-configured exploratory questions to steer the LLM's code generation process toward a specific, logical fix.
\item \textbf{The Sanitizer (Code Reviewer):} Ensures the reliability of the LLM's output. It cleans and formats the generated text to produce a pure, syntactically correct Python file, removing conversational artifacts or inconsistencies before the code is passed to the Evaluator.
\end{itemize}

\textbf{Iterative Workflow:} The Auto-RCA system operates in a closed loop designed for progressive refinement:
\begin{enumerate}
\item \textbf{Baseline Testing (Round 0):} An initial code version is run on the benchmark, establishing a baseline F1-Score and generating the first set of Bad Cases.
\item \textbf{Failure Analysis:} The Orchestrator passes the Bad Cases to the Bad Case Analyzer, which identifies the most critical systemic flaw (e.g., a consistent pattern of generating "EXTRA\_ROOT\_CAUSE").
\item \textbf{Guided Code Generation:} The LLM Agent uses this analysis to construct a targeted prompt and queries the LLM for a new, improved version of the solution code.
\item \textbf{Evaluation \& Decision:} The candidate code is sanitized and re-evaluated on the benchmark. The Orchestrator compares the new F1-score to the previous best. If the score improves, the new code is accepted and becomes the basis for the next iteration. If not, it is rejected.
\item \textbf{Iteration:} The process repeats, using the new set of (fewer) Bad Cases from the successful run as input for the next round of analysis, progressively refining the solution.
\end{enumerate}

\subsection{Evaluation Framework and Metrics}

To systematically measure LLM performance on TN-RCA530, we developed a comprehensive evaluation framework and a set of precise metrics.

\textbf{Evaluation Protocol:} Our framework is designed for automated execution. Given a model's API endpoint, the framework presents each of the 530 scenarios as an input knowledge graph. The model is prompted with a structured set of instructions to analyze the graph and return its findings in a standardized JSON format. This format requires the model to identify the target alarms and, for each, list the inferred root causes, including a textual description ($cause\_description$), the associated equipment ID ($equipment\_id$), and a proposed solution.

\begin{table*}[!h]
\centering
\caption{Performance Breakdown by Scenario Difficulty on TN-RCA530. The best-performing value in each column is highlighted in \textbf{bold}.}
\label{tab:performance_by_difficulty}
\renewcommand{\arraystretch}{1.4}
\sisetup{
  detect-weight=true,
  detect-family=true,
  table-format=1.4
}
\begin{tabular*}{\textwidth}{@{\extracolsep{\fill}} l *{9}{S} @{}}
\toprule
& \multicolumn{3}{c}{\textbf{Precision@1}} & \multicolumn{3}{c}{\textbf{Recall@1}} & \multicolumn{3}{c}{\textbf{F1-Score@1}} \\
\cmidrule(lr){2-4} \cmidrule(lr){5-7} \cmidrule(lr){8-10}
\textbf{Model} & {\textbf{Simple}} & {\textbf{Difficult}} & {\textbf{Mixed}} & {\textbf{Simple}} & {\textbf{Difficult}} & {\textbf{Mixed}} & {\textbf{Simple}} & {\textbf{Difficult}} & {\textbf{Mixed}} \\
\midrule
DeepSeek-R1-671B   & 0.8448          & 0.4240          & 0.4471          & 0.8448          & 0.9600          & 0.9537          & 0.8448          & 0.5749          & 0.5897 \\
Gemini-2.5-Pro     & \textbf{0.8621} & 0.4311          & 0.4356          & \textbf{0.8621} & 0.9658          & 0.9134          & \textbf{0.8621} & 0.5824          & 0.5899 \\
Claude-3.5-Sonnet & \textbf{0.8621} & 0.4021          & 0.4234          & \textbf{0.8621} & 0.9653          & 0.9537          & \textbf{0.8621} & 0.5677          & 0.5934 \\
Claude-3.7-Sonnet & \textbf{0.8621} & 0.4370          & 0.4441          & \textbf{0.8621} & 0.9800 & 0.9626          & \textbf{0.8621} & 0.5912          & 0.6078 \\
Claude-Sonnet-4   & \textbf{0.8621} & 0.4323 & 0.4561          & \textbf{0.8621} & 0.9783          & 0.9586          & \textbf{0.8621} & 0.5995 & 0.6181 \\
Qwen3-4B           & 0.6752          & 0.3680          & 0.3782 & 0.7040          & 0.8302          & 0.8038          & 0.6741          & 0.4955          & 0.5144 \\
Qwen3-32B          & 0.8302          & 0.4277          & 0.4498          & 0.8563          & 0.9756          & 0.9691         & 0.8311          & 0.5789          & 0.6142 \\
Qwen-QWQ           & \textbf{0.8621} & \textbf{0.4381 }         & \textbf{0.4613}          & \textbf{0.8621} & 0.9744          & 0.9682          & \textbf{0.8621} & 0.5900          & 0.6248 \\
Qwen3-235B         & \textbf{0.8621} & 0.4374          & 0.4607          & \textbf{0.8621} & \textbf{0.9810  }        & \textbf{0.9736} & \textbf{0.8621} & \textbf{0.6049 }         & \textbf{0.6254} \\
\bottomrule
\end{tabular*}
\end{table*}

\textbf{Primary Metric (F1-Score):} Given the multi-label nature of the task (i.e., multiple alarms may be present, and a single alarm could have multiple root causes), we adopt the F1-Score as our primary evaluation metric. It provides a balanced measure of a model's ability to correctly identify all true root causes without introducing false ones.

The F1-Score is calculated as follows:
\begin{itemize}[leftmargin=*]
    \item \textbf{Extraction:} The framework parses the model's generated JSON to extract the set of predicted root causes ($P$). It also extracts the ground-truth set of root causes ($T$) from the corresponding label file. Each root cause is treated as a tuple of ($cause\_description$, $equipment\_id$).

    \item \textbf{Matching: }A predicted root cause $p \in P$ is considered a True Positive (TP) if it exactly matches a ground-truth root cause $t \in T$.

    \item \textbf{Calculation:} We compute Precision and Recall: $$ \text{Precision} = \frac{|P \cap T|}{|P|} $$ $$ \text{Recall} = \frac{|P \cap T|}{|T|} $$

    \item The F1-Score is the harmonic mean of Precision and Recall: $$ F_1 = 2 \times \frac{\text{Precision} \times \text{Recall}}{\text{Precision} + \text{Recall}} $$

\end{itemize}

The final reported score for a model is the macro-average of the F1-Scores across all 530 scenarios. Our framework also reports overall Precision and Recall to provide deeper insights into model behavior.

%% file: experiments.tex
\section{Experiments}

This section presents a series of experiments designed to answer two core research questions:
\begin{enumerate}
    \item What is the baseline performance of current state-of-the-art Large Language Models (LLMs) when applied directly to the TN-RCA530 benchmark?
    \item Can the Auto-RCA agentic framework effectively improve the performance of solving the problems within TN-RCA530?
\end{enumerate}

\subsection{Experimental Setup}
\begin{itemize}
    \item \textbf{Benchmark:} All experiments are conducted on the complete TN-RCA530 benchmark, which comprises 530 real-world fault scenarios grounded in expert-verified knowledge graphs. This benchmark is further categorized into 29 "Simple" scenarios and 501 "Difficult" scenarios, with "Mixed" representing the overall dataset.
    \item \textbf{Evaluation Metrics:} The primary evaluation metric is the macro-averaged F1-score, calculated across all 530 scenarios. To provide a more comprehensive performance view, we also report Precision@k (P@k) and Recall@k (R@k), where k=1.
    \item \textbf{Baseline Models:} To establish a robust baseline, we evaluated a diverse suite of prominent closed-source and open-source LLMs. As shown in Table 1, this includes models from the DeepSeek and Qwen series, as well as Google's Gemini, and Anthropic's Claude families.
    \item \textbf{Agent Configuration:} For the Auto-RCA system, we selected several leading models that support the long context windows required by the framework (at least 64K tokens). Each experimental run consists of five iterative optimization rounds, starting from the same initial code base.
\end{itemize}

\subsection{RQ1: Baseline Performance of LLMs on TN-RCA530}
To answer our first research question, we measured the "out-of-the-box" performance of various LLMs when directly applied to the TN-RCA530 benchmark. This experiment assesses the intrinsic reasoning capabilities of these models when tasked with solving the RCA problems directly, without the aid of an iterative refinement framework. The TN-RCA530 benchmark is particularly challenging, comprising 530 real-world fault scenarios, with a significant majority (501 "Difficult" scenarios) requiring complex reasoning, alongside a smaller set of 29 "Simple" scenarios.

The results, summarized in Table 1, are revealing. For "Simple" scenarios, several models, including Gemini-2.5-Pro, Claude-3.5-Sonnet, Claude-3.7-Sonnet, Claude-Sonnet-4, Qwen-QWQ, and Qwen3-235B, achieve an identical and high Precision@1, Recall@1, and F1-Score@1 of \textbf{0.8621}. This indicates that for less complex fault diagnosis tasks, many state-of-the-art LLMs demonstrate comparable and strong performance.

\begin{figure}[htbp]
\centering
\includegraphics[width=\columnwidth]{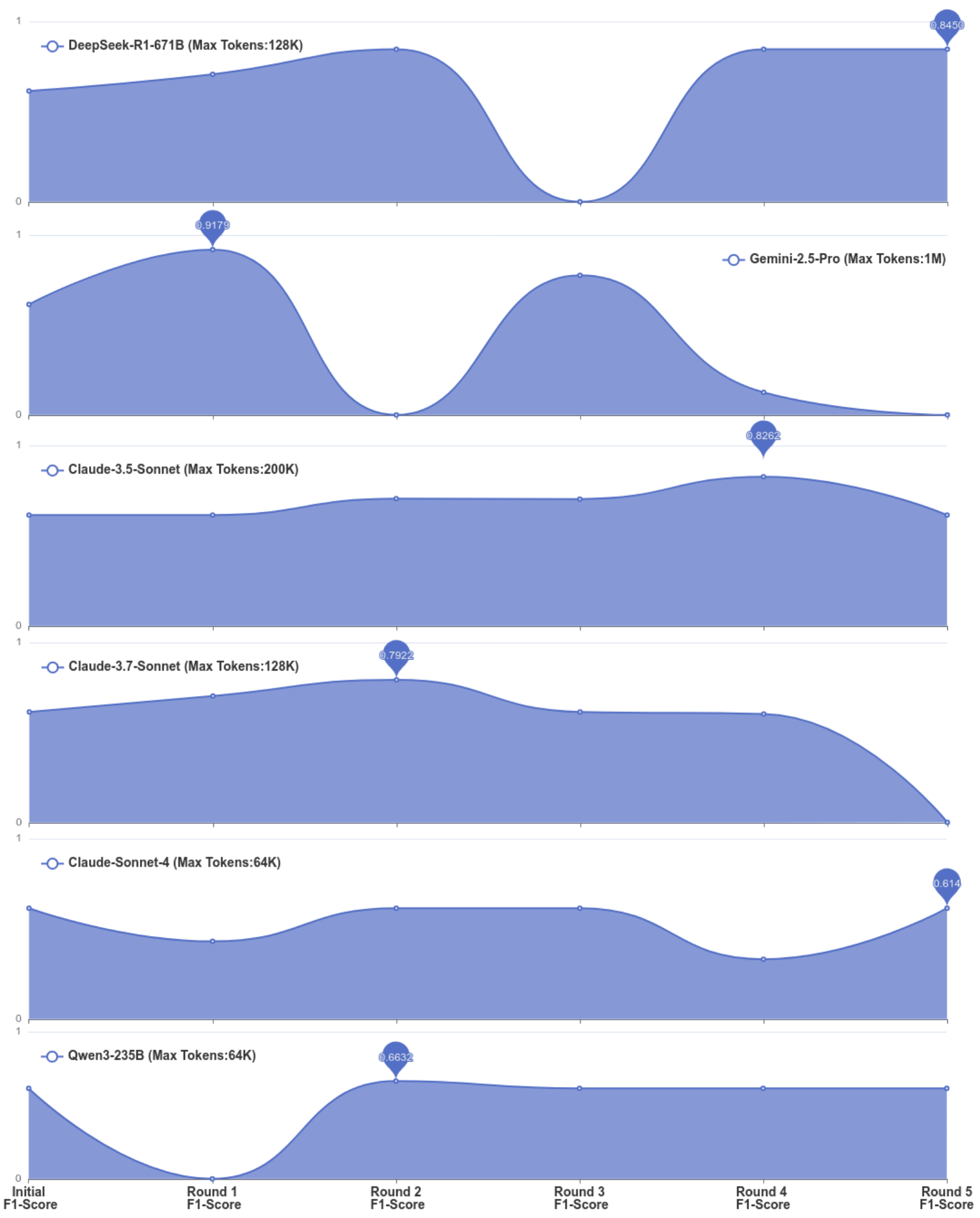} 
\caption{Performance Improvement on TN-RCA530 using the Auto-RCA Framework. The line chart shows the F1-score evolution of the best solution code across five refinement iterations. Markpoints indicate the highest F1-score achieved by each model on the mixed dataset. A score of 0.0000 signifies that a proposed code modification was rejected due to a performance decrease or an error, ensuring monotonic improvement.}
\label{auto_rca}
\end{figure}

\begin{table*}[!h]
\centering
\caption{Performance Breakdown of Best Solution Code by Scenario Difficulty. The best-performing value in each column is highlighted in \textbf{bold}.}
\label{tab:performance_best_solution}
\renewcommand{\arraystretch}{1.4}
\sisetup{
  detect-weight=true,
  detect-family=true,
  table-format=1.4
}
\begin{tabular*}{\textwidth}{@{\extracolsep{\fill}} l *{9}{S} @{}}
\toprule
& \multicolumn{3}{c}{\textbf{Precision@1}} & \multicolumn{3}{c}{\textbf{Recall@1}} & \multicolumn{3}{c}{\textbf{F1-Score@1}} \\
\cmidrule(lr){2-4} \cmidrule(lr){5-7} \cmidrule(lr){8-10}
\textbf{Model} & {\textbf{Simple}} & {\textbf{Difficult}} & {\textbf{Mixed}} & {\textbf{Simple}} & {\textbf{Difficult}} & {\textbf{Mixed}} & {\textbf{Simple}} & {\textbf{Difficult}} & {\textbf{Mixed}} \\
\midrule
DeepSeek-R1-671B   & 0.7832          & 0.8386          & 0.8115          & 0.8409          & 0.9627          & 0.9288          & 0.7997          & 0.8746          & 0.8450 \\
Gemini-2.5-Pro     & \textbf{0.9483} & \textbf{0.8920} & \textbf{0.8951} & \textbf{0.9828} & 0.9763          & 0.9767          & \textbf{0.9540} & \textbf{0.9158} & \textbf{0.9179} \\
Claude-3.5-Sonnet & 0.9234          & 0.8547          & 0.8097          & 0.7453          & 0.9113          & 0.8633          & 0.8250          & 0.8721          & 0.8262 \\
Claude-3.7-Sonnet & 0.9310          & 0.8500          & 0.8544          & 0.7586          & 0.7660          & 0.7656          & 0.8161          & 0.7908          & 0.7922 \\
Claude-Sonnet-4   & \textbf{0.9483} & 0.4407          & 0.4685          & \textbf{0.9828} & \textbf{0.9823} & \textbf{0.9823} & \textbf{0.9540} & 0.5943          & 0.6140 \\
Qwen3-235B         & 0.9109          & 0.6950          & 0.6682          & 0.7839          & 0.6930          & 0.6626          & 0.8422          & 0.6923          & 0.6632 \\
\bottomrule
\end{tabular*}
\end{table*}

However, the performance significantly drops when models are faced with "Difficult" scenarios. In this category, Qwen3-235B exhibits the best Recall@1 (\textbf{0.9810}) and F1-Score@1 (\textbf{0.6049}), while Qwen-QWQ achieves the highest Precision@1 (\textbf{0.4381}). The "Mixed" scenario, which reflects the overall dataset composition, shows Qwen-QWQ leading in Precision@1 (\textbf{0.4613}), and Qwen3-235B leading in both Recall@1 (\textbf{0.9736}) and F1-Score@1 (\textbf{0.6254}). Notably, the Qwen series models generally demonstrate strong performance across the "Mixed" and "Difficult" categories. This can be attributed to the fact that the fault descriptions in our benchmark are primarily in Chinese, where Qwen models, being developed with a strong focus on Chinese language capabilities, may possess a distinct advantage in understanding and processing the nuances of the problem statements.

The observed performance ceiling, with even the best F1-Score@1 in "Mixed" scenarios reaching only \textbf{0.6254} (Qwen3-235B), underscores the inherent difficulty of the TN-RCA530 benchmark for direct LLM application. It confirms that specialized, graph-based reasoning in a complex domain remains a significant challenge for general-purpose LLMs, and that merely scaling models is insufficient for mastery without further refinement or an agentic approach.

\subsection{RQ2: Effectiveness of the Auto-RCA Framework}
To answer our second research question, we evaluated the effectiveness of the Auto-RCA framework. This experiment does not measure an improvement in the LLM's underlying intelligence but rather the performance of the entire problem-solving system, which uses an LLM as a code-generating component. The goal is to determine if our agentic, iterative refinement process can produce a final solution with higher accuracy on the benchmark.

Figure~\ref{auto_rca} shows the performance of the final solution code as it is iteratively refined by the Auto-RCA framework over five rounds. The results demonstrate a dramatic improvement in problem-solving capability compared to the baseline direct application of LLMs. Notably, when using Gemini-2.5-Pro as the reasoning engine within the framework, the solution's F1-score is elevated from an initial baseline of 0.5899 to an impressive \textbf{0.9179} in Round 1, which stands as the best overall performance achieved by any model in the framework. This substantial increase highlights the framework's ability to significantly enhance problem-solving accuracy.

The iterative progress is evident across the rounds for various models. A score of 0.0000 indicates a round where the generated code was syntactically invalid or performed worse than the previous best, leading the Orchestrator to reject the change—a key feature of the system's robustness that prevents performance regression and ensures only improvements are adopted.

A critical factor contributing to Gemini-2.5-Pro's superior performance within the Auto-RCA framework is its significantly larger Max Tokens capacity (1M). This extensive context window allows Gemini-2.5-Pro to process a greater volume of effective information, including detailed problem descriptions, historical interactions, and intermediate reasoning steps, without truncation. In contrast, other models with smaller context windows (e.g., Claude-Sonnet-4 and Qwen3-235B, both at 64K tokens) are more susceptible to information loss due to truncation, which can hinder their ability to generate optimal code modifications and lead to suboptimal performance. This suggests that for complex, iterative code generation tasks, the ability to maintain a comprehensive context is paramount.

To further analyze the framework's output, we assessed the best-performing solution code (from Gemini-2.5-Pro, F1-score 0.9179) across different scenario difficulties, as detailed in Table~\ref{tab:performance_best_solution}.

\begin{itemize}
    \item \textbf{Simple Scenarios:} The solution achieved a near-perfect F1-score of 0.9540, demonstrating mastery over foundational problems with clear causal links.

    \item \textbf{Difficult Scenarios:} Performance dropped to an F1-score of 0.9158, revealing the current limits of the solution when faced with complex causal chains or ambiguous signals. The higher Recall (0.9763) compared to Precision (0.8920) suggests the model tends to identify most true positives but also flags some false positives, adopting an "aggressive" diagnostic strategy in complex cases.

    \item \textbf{Mixed Scenarios:} The F1-score of 0.9179 aligns closely with the overall iterative performance, confirming the benchmark's balanced composition and the solution's real-world effectiveness.

\end{itemize}

In summary, the Auto-RCA framework is a highly effective paradigm for complex reasoning tasks. It achieves a final F1-score of 0.9179 (Gemini-2.5-Pro), a substantial increase over the best baseline F1-Score@1 of 0.6254 (Qwen3-235B in Mixed scenarios). This validates our hypothesis that an autonomous, self-optimizing agentic architecture can effectively master domain-specific challenges like those in TN-RCA530.

%% file: conclusion.tex
\section{Conclusion}
This paper makes two core contributions to AI-driven network management. First, we introduced TN-RCA530, a public, real-world benchmark that provides a standardized platform to measure and drive progress in telecom RCA. Second, we proposed Auto-RCA, a novel agentic framework that demonstrates a clear path to mastering this benchmark.

The central insight of our work is that for complex, domain-specific reasoning tasks, the most effective paradigm may not be the direct application of LLMs. Instead, superior performance can be achieved by leveraging them as components within autonomous agentic frameworks that iteratively refine verifiable solutions. Future work will focus on extending the benchmark with greater complexity and generalizing the Auto-RCA framework to other challenging, graph-based problem domains.

%% file: anonymous-submission-latex-2025.bbl
\begin{thebibliography}{32}
\providecommand{\natexlab}[1]{#1}

\bibitem[{Amayuelas et~al.(2025)Amayuelas, Sain, Kaur, and Smiley}]{amayuelas2025groundingllmreasoningknowledge}
Amayuelas, A.; Sain, J.; Kaur, S.; and Smiley, C. 2025.
\newblock Grounding LLM Reasoning with Knowledge Graphs.
\newblock arXiv:2502.13247.

\bibitem[{Arora et~al.(2024)Arora, Sonwane, Wadhwa, Mehrotra, Utpala, Bairi, Kanade, and Natarajan}]{arora2024masaimodulararchitecturesoftwareengineering}
Arora, D.; Sonwane, A.; Wadhwa, N.; Mehrotra, A.; Utpala, S.; Bairi, R.; Kanade, A.; and Natarajan, N. 2024.
\newblock MASAI: Modular Architecture for Software-engineering AI Agents.
\newblock arXiv:2406.11638.

\bibitem[{Barriah et~al.(2025)Barriah, De~Domenico, Powell, Sana, Wang, Piovesan, and Debbah}]{otellm}
Barriah, L.; De~Domenico, A.; Powell, L.; Sana, M.; Wang, P.; Piovesan, N.; and Debbah, M. 2025.
\newblock GSMA Open-Telco LLM Benchmarks.
\newblock \url{https://huggingface.co/blog/otellm/gsma-benchmarks}.

\bibitem[{Bouloutas, Calo, and Finkel(1994)}]{acafiicn}
Bouloutas, A.; Calo, S.; and Finkel, A. 1994.
\newblock Alarm correlation and fault identification in communication networks.
\newblock \emph{Communications, IEEE Transactions on}, 42: 523 -- 533.

\bibitem[{Campos et~al.(2025)Campos, Shariffdeen, Ulges, and Noller}]{campos2025empiricalevaluationgeneralizableautomated}
Campos, V.; Shariffdeen, R.; Ulges, A.; and Noller, Y. 2025.
\newblock Empirical Evaluation of Generalizable Automated Program Repair with Large Language Models.
\newblock arXiv:2506.03283.

\bibitem[{DeepSeek-AI et~al.(2025)DeepSeek-AI, Guo, Yang, Zhang, Song, Zhang, Xu, Zhu, Ma, Wang, Bi, Zhang, Yu, Wu, Wu, Gou, Shao, Li, Gao, Liu, Xue, Wang, Wu, Feng, Lu, Zhao, Deng, Zhang, Ruan, Dai, Chen, Ji, Li, Lin, Dai, Luo, Hao, Chen, Li, Zhang, Bao, Xu, Wang, Ding, Xin, Gao, Qu, Li, Guo, Li, Wang, Chen, Yuan, Qiu, Li, Cai, Ni, Liang, Chen, Dong, Hu, Gao, Guan, Huang, Yu, Wang, Zhang, Zhao, Wang, Zhang, Xu, Xia, Zhang, Zhang, Tang, Li, Wang, Li, Tian, Huang, Zhang, Wang, Chen, Du, Ge, Zhang, Pan, Wang, Chen, Jin, Chen, Lu, Zhou, Chen, Ye, Wang, Yu, Zhou, Pan, Li, Zhou, Wu, Ye, Yun, Pei, Sun, Wang, Zeng, Zhao, Liu, Liang, Gao, Yu, Zhang, Xiao, An, Liu, Wang, Chen, Nie, Cheng, Liu, Xie, Liu, Yang, Li, Su, Lin, Li, Jin, Shen, Chen, Sun, Wang, Song, Zhou, Wang, Shan, Li, Wang, Wei, Zhang, Xu, Li, Zhao, Sun, Wang, Yu, Zhang, Shi, Xiong, He, Piao, Wang, Tan, Ma, Liu, Guo, Ou, Wang, Gong, Zou, He, Xiong, Luo, You, Liu, Zhou, Zhu, Xu, Huang, Li, Zheng, Zhu, Ma, Tang, Zha, Yan, Ren, Ren, Sha, Fu, Xu, Xie, Zhang,
  Hao, Ma, Yan, Wu, Gu, Zhu, Liu, Li, Xie, Song, Pan, Huang, Xu, Zhang, and Zhang}]{deepseekai2025deepseekr1incentivizingreasoningcapability}
DeepSeek-AI; Guo, D.; Yang, D.; Zhang, H.; Song, J.; Zhang, R.; Xu, R.; Zhu, Q.; Ma, S.; Wang, P.; Bi, X.; Zhang, X.; Yu, X.; Wu, Y.; Wu, Z.~F.; Gou, Z.; Shao, Z.; Li, Z.; Gao, Z.; Liu, A.; Xue, B.; Wang, B.; Wu, B.; Feng, B.; Lu, C.; Zhao, C.; Deng, C.; Zhang, C.; Ruan, C.; Dai, D.; Chen, D.; Ji, D.; Li, E.; Lin, F.; Dai, F.; Luo, F.; Hao, G.; Chen, G.; Li, G.; Zhang, H.; Bao, H.; Xu, H.; Wang, H.; Ding, H.; Xin, H.; Gao, H.; Qu, H.; Li, H.; Guo, J.; Li, J.; Wang, J.; Chen, J.; Yuan, J.; Qiu, J.; Li, J.; Cai, J.~L.; Ni, J.; Liang, J.; Chen, J.; Dong, K.; Hu, K.; Gao, K.; Guan, K.; Huang, K.; Yu, K.; Wang, L.; Zhang, L.; Zhao, L.; Wang, L.; Zhang, L.; Xu, L.; Xia, L.; Zhang, M.; Zhang, M.; Tang, M.; Li, M.; Wang, M.; Li, M.; Tian, N.; Huang, P.; Zhang, P.; Wang, Q.; Chen, Q.; Du, Q.; Ge, R.; Zhang, R.; Pan, R.; Wang, R.; Chen, R.~J.; Jin, R.~L.; Chen, R.; Lu, S.; Zhou, S.; Chen, S.; Ye, S.; Wang, S.; Yu, S.; Zhou, S.; Pan, S.; Li, S.~S.; Zhou, S.; Wu, S.; Ye, S.; Yun, T.; Pei, T.; Sun, T.; Wang, T.; Zeng, W.;
  Zhao, W.; Liu, W.; Liang, W.; Gao, W.; Yu, W.; Zhang, W.; Xiao, W.~L.; An, W.; Liu, X.; Wang, X.; Chen, X.; Nie, X.; Cheng, X.; Liu, X.; Xie, X.; Liu, X.; Yang, X.; Li, X.; Su, X.; Lin, X.; Li, X.~Q.; Jin, X.; Shen, X.; Chen, X.; Sun, X.; Wang, X.; Song, X.; Zhou, X.; Wang, X.; Shan, X.; Li, Y.~K.; Wang, Y.~Q.; Wei, Y.~X.; Zhang, Y.; Xu, Y.; Li, Y.; Zhao, Y.; Sun, Y.; Wang, Y.; Yu, Y.; Zhang, Y.; Shi, Y.; Xiong, Y.; He, Y.; Piao, Y.; Wang, Y.; Tan, Y.; Ma, Y.; Liu, Y.; Guo, Y.; Ou, Y.; Wang, Y.; Gong, Y.; Zou, Y.; He, Y.; Xiong, Y.; Luo, Y.; You, Y.; Liu, Y.; Zhou, Y.; Zhu, Y.~X.; Xu, Y.; Huang, Y.; Li, Y.; Zheng, Y.; Zhu, Y.; Ma, Y.; Tang, Y.; Zha, Y.; Yan, Y.; Ren, Z.~Z.; Ren, Z.; Sha, Z.; Fu, Z.; Xu, Z.; Xie, Z.; Zhang, Z.; Hao, Z.; Ma, Z.; Yan, Z.; Wu, Z.; Gu, Z.; Zhu, Z.; Liu, Z.; Li, Z.; Xie, Z.; Song, Z.; Pan, Z.; Huang, Z.; Xu, Z.; Zhang, Z.; and Zhang, Z. 2025.
\newblock DeepSeek-R1: Incentivizing Reasoning Capability in LLMs via Reinforcement Learning.
\newblock arXiv:2501.12948.

\bibitem[{Dehal, Sharma, and Rajabi(2025)}]{make7020038}
Dehal, R.~S.; Sharma, M.; and Rajabi, E. 2025.
\newblock Knowledge Graphs and Their Reciprocal Relationship with Large Language Models.
\newblock \emph{Machine Learning and Knowledge Extraction}, 7(2).

\bibitem[{Ethiraj et~al.(2025)Ethiraj, Vijay, Menon, and Berscilla}]{ethiraj2025efficienttelecomspecificllm}
Ethiraj, V.; Vijay, D.; Menon, S.; and Berscilla, H. 2025.
\newblock Efficient Telecom Specific LLM: TSLAM-Mini with QLoRA and Digital Twin Data.
\newblock arXiv:2505.07877.

\bibitem[{Ibrahim et~al.(2024)Ibrahim, Aboulela, Ibrahim, and Kashef}]{Ibrahim2024survey}
Ibrahim, N.; Aboulela, S.; Ibrahim, A.; and Kashef, R. 2024.
\newblock A survey on augmenting knowledge graphs (KGs) with large language models (LLMs): models, evaluation metrics, benchmarks, and challenges.
\newblock \emph{Discover Artificial Intelligence}, 4(1).

\bibitem[{Jakobson and Weissman(1995)}]{rttnm}
Jakobson, G.; and Weissman, M. 1995.
\newblock Real-time telecommunication network management: extending event correlation with temporal constraints.
\newblock In \emph{Proceedings of the Fourth International Symposium on Integrated Network Management IV}, 290–301. GBR: Chapman \& Hall, Ltd.
\newblock ISBN 0412715708.

\bibitem[{Ji et~al.(2024)Ji, Liu, Xi, Zhang, and Li}]{electronics13245037}
Ji, S.; Liu, L.; Xi, J.; Zhang, X.; and Li, X. 2024.
\newblock KLR-KGC: Knowledge-Guided LLM Reasoning for Knowledge Graph Completion.
\newblock \emph{Electronics}, 13(24).

\bibitem[{Just, Jalali, and Ernst(2014)}]{Just2014Defects4J}
Just, R.; Jalali, D.; and Ernst, M.~D. 2014.
\newblock {Defects4J}: A Database of Existing Faults to Enable Controlled Testing Studies for Java Programs.
\newblock In \emph{Proceedings of the 2014 International Symposium on Software Testing and Analysis}, ISSTA '14, 437--440. New York, NY, USA: ACM.

\bibitem[{Kamoi et~al.(2024)Kamoi, Zhang, Zhang, Han, and Zhang}]{kamoi-etal-2024-llms}
Kamoi, R.; Zhang, Y.; Zhang, N.; Han, J.; and Zhang, R. 2024.
\newblock When Can {LLM}s Actually Correct Their Own Mistakes? A Critical Survey of Self-Correction of {LLM}s.
\newblock \emph{Transactions of the Association for Computational Linguistics}, 12: 1417--1440.

\bibitem[{Kong et~al.(2024)Kong, Cheng, Xie, Liu, Du, and Guo}]{kong2024contrastrepairenhancingconversationbasedautomated}
Kong, J.; Cheng, M.; Xie, X.; Liu, S.; Du, X.; and Guo, Q. 2024.
\newblock ContrastRepair: Enhancing Conversation-Based Automated Program Repair via Contrastive Test Case Pairs.
\newblock arXiv:2403.01971.

\bibitem[{Li, Yang, and Chen(2023)}]{ararci}
Li, M.; Yang, M.; and Chen, P. 2023.
\newblock Alarm reduction and root cause inference based on association mining in communication network.
\newblock \emph{Frontiers in Computer Science}, Volume 5 - 2023.

\bibitem[{Liang, Liu, and Liu(2020)}]{Liang2020ABN}
Liang, R.; Liu, F.; and Liu, J. 2020.
\newblock A Belief Network Reasoning Framework for Fault Localization in Communication Networks.
\newblock \emph{Sensors (Basel, Switzerland)}, 20.

\bibitem[{Ma et~al.(2024)Ma, Zhen, Ren, Zhang, Zhang, and Dong}]{en17225758}
Ma, X.; Zhen, W.; Ren, H.; Zhang, G.; Zhang, K.; and Dong, H. 2024.
\newblock A Method for Fault Localization in Distribution Networks with High Proportions of Distributed Generation Based on Graph Convolutional Networks.
\newblock \emph{Energies}, 17(22).

\bibitem[{Ma et~al.(2025)Ma, Li, Dong, Jiang, Cao, Chen, Huang, and Li}]{ma2025thinkinglongerlargerenhancing}
Ma, Y.; Li, Y.; Dong, Y.; Jiang, X.; Cao, R.; Chen, J.; Huang, F.; and Li, B. 2025.
\newblock Thinking Longer, Not Larger: Enhancing Software Engineering Agents via Scaling Test-Time Compute.
\newblock arXiv:2503.23803.

\bibitem[{Maatouk et~al.(2025)Maatouk, Ampudia, Ying, and Tassiulas}]{maatouk2025telellmsseriesspecializedlarge}
Maatouk, A.; Ampudia, K.~C.; Ying, R.; and Tassiulas, L. 2025.
\newblock Tele-LLMs: A Series of Specialized Large Language Models for Telecommunications.
\newblock arXiv:2409.05314.

\bibitem[{Pan et~al.(2024)Pan, Saxon, Xu, Nathani, Wang, and Wang}]{pan-etal-2024-automatically}
Pan, L.; Saxon, M.; Xu, W.; Nathani, D.; Wang, X.; and Wang, W.~Y. 2024.
\newblock Automatically Correcting Large Language Models: Surveying the Landscape of Diverse Automated Correction Strategies.
\newblock \emph{Transactions of the Association for Computational Linguistics}, 12: 484--506.

\bibitem[{Pham et~al.(2025)Pham, Zhang, Ha, Salim, and Zhang}]{pham2025rcaevalbenchmarkrootcause}
Pham, L.; Zhang, H.; Ha, H.; Salim, F.; and Zhang, X. 2025.
\newblock RCAEval: A Benchmark for Root Cause Analysis of Microservice Systems with Telemetry Data.
\newblock arXiv:2412.17015.

\bibitem[{Roy et~al.(2024)Roy, Zhang, Bhave, Bansal, Las-Casas, Fonseca, and Rajmohan}]{elafrca}
Roy, D.; Zhang, X.; Bhave, R.; Bansal, C.; Las-Casas, P.; Fonseca, R.; and Rajmohan, S. 2024.
\newblock Exploring LLM-Based Agents for Root Cause Analysis.
\newblock In \emph{Companion Proceedings of the 32nd ACM International Conference on the Foundations of Software Engineering}, FSE 2024, 208–219. New York, NY, USA: Association for Computing Machinery.
\newblock ISBN 9798400706585.

\bibitem[{Shen et~al.(2025)Shen, Wang, Zhang, and Cambria}]{shenetal}
Shen, T.; Wang, J.; Zhang, X.; and Cambria, E. 2025.
\newblock Reasoning with Trees: Faithful Question Answering over Knowledge Graph.
\newblock In Rambow, O.; Wanner, L.; Apidianaki, M.; Al-Khalifa, H.; Eugenio, B.~D.; and Schockaert, S., eds., \emph{Proceedings of the 31st International Conference on Computational Linguistics}, 3138--3157. Abu Dhabi, UAE: Association for Computational Linguistics.

\bibitem[{Shen, Wang, and Xia(2025)}]{shen2025reasonalignrespondaligningllmreasoning}
Shen, X.; Wang, F.; and Xia, R. 2025.
\newblock Reason-Align-Respond: Aligning LLM Reasoning with Knowledge Graphs for KGQA.
\newblock arXiv:2505.20971.

\bibitem[{Team et~al.(2025)Team, Du, Gao, Xing, Jiang, Chen, Li, Xiao, Du, Liao, Tang, Wang, Zhang, Yuan, Lu, Tang, Sung, Wei, Lai, Guo, Zhu, Ding, Hu, Yang, Zhang, Yao, Zhao, Lu, Li, Yu, Gao, Zheng, Yuan, Chen, Guo, Su, Wang, Zhao, Zhang, Liu, Yan, Wu, Shi, Ye, Yu, Dong, Zhang, Ma, Pan, Gong, Liu, Ma, Wei, Cao, Huang, Jiang, Gao, Xiong, He, Huang, Xu, Wu, He, Wei, Jia, Wu, Xu, Zu, Zhou, Pan, Charles, Li, Hu, Liu, Chen, Wang, Liu, Qin, Liu, Yang, Bao, Du, Wu, Wang, Zhou, Wang, Li, Zhu, Zhang, Wang, Yang, Huang, Huang, Xu, Yang, and Lin}]{kimiteam2025kimik15scalingreinforcement}
Team, K.; Du, A.; Gao, B.; Xing, B.; Jiang, C.; Chen, C.; Li, C.; Xiao, C.; Du, C.; Liao, C.; Tang, C.; Wang, C.; Zhang, D.; Yuan, E.; Lu, E.; Tang, F.; Sung, F.; Wei, G.; Lai, G.; Guo, H.; Zhu, H.; Ding, H.; Hu, H.; Yang, H.; Zhang, H.; Yao, H.; Zhao, H.; Lu, H.; Li, H.; Yu, H.; Gao, H.; Zheng, H.; Yuan, H.; Chen, J.; Guo, J.; Su, J.; Wang, J.; Zhao, J.; Zhang, J.; Liu, J.; Yan, J.; Wu, J.; Shi, L.; Ye, L.; Yu, L.; Dong, M.; Zhang, N.; Ma, N.; Pan, Q.; Gong, Q.; Liu, S.; Ma, S.; Wei, S.; Cao, S.; Huang, S.; Jiang, T.; Gao, W.; Xiong, W.; He, W.; Huang, W.; Xu, W.; Wu, W.; He, W.; Wei, X.; Jia, X.; Wu, X.; Xu, X.; Zu, X.; Zhou, X.; Pan, X.; Charles, Y.; Li, Y.; Hu, Y.; Liu, Y.; Chen, Y.; Wang, Y.; Liu, Y.; Qin, Y.; Liu, Y.; Yang, Y.; Bao, Y.; Du, Y.; Wu, Y.; Wang, Y.; Zhou, Z.; Wang, Z.; Li, Z.; Zhu, Z.; Zhang, Z.; Wang, Z.; Yang, Z.; Huang, Z.; Huang, Z.; Xu, Z.; Yang, Z.; and Lin, Z. 2025.
\newblock Kimi k1.5: Scaling Reinforcement Learning with LLMs.
\newblock arXiv:2501.12599.

\bibitem[{Team(2025)}]{qwq32b}
Team, Q. 2025.
\newblock QwQ-32B: Embracing the Power of Reinforcement Learning.

\bibitem[{Wang et~al.(2025)Wang, Zhang, Li, Yuan, Xiao, Zhuang, and Yu}]{wang2025tamofinegrainedrootcauseanalysis}
Wang, Q.; Zhang, X.; Li, M.; Yuan, Y.; Xiao, M.; Zhuang, F.; and Yu, D. 2025.
\newblock TAMO:Fine-Grained Root Cause Analysis via Tool-Assisted LLM Agent with Multi-Modality Observation Data in Cloud-Native Systems.
\newblock arXiv:2504.20462.

\bibitem[{Wang et~al.(2024)Wang, Huang, Zhou, Chen, and Wang}]{pr12112312}
Wang, Z.; Huang, B.; Zhou, B.; Chen, J.; and Wang, Y. 2024.
\newblock An Enhanced Fault Localization Technique for Distribution Networks Utilizing Cost-Sensitive Graph Neural Networks.
\newblock \emph{Processes}, 12(11).

\bibitem[{Xu et~al.(2025)Xu, Zhang, Zhong, He, Zhang, Lin, Pei, He, Zhang, and Zhang}]{xu2025openrca}
Xu, J.; Zhang, Q.; Zhong, Z.; He, S.; Zhang, C.; Lin, Q.; Pei, D.; He, P.; Zhang, D.; and Zhang, Q. 2025.
\newblock OpenRCA: Can Large Language Models Locate the Root Cause of Software Failures?
\newblock In \emph{The Thirteenth International Conference on Learning Representations}.

\bibitem[{Yang et~al.(2025)Yang, Li, Yang, Zhang, Hui, Zheng, Yu, Gao, Huang, Lv, Zheng, Liu, Zhou, Huang, Hu, Ge, Wei, Lin, Tang, Yang, Tu, Zhang, Yang, Yang, Zhou, Zhou, Lin, Dang, Bao, Yang, Yu, Deng, Li, Xue, Li, Zhang, Wang, Zhu, Men, Gao, Liu, Luo, Li, Tang, Yin, Ren, Wang, Zhang, Ren, Fan, Su, Zhang, Zhang, Wan, Liu, Wang, Cui, Zhang, Zhou, and Qiu}]{yang2025qwen3technicalreport}
Yang, A.; Li, A.; Yang, B.; Zhang, B.; Hui, B.; Zheng, B.; Yu, B.; Gao, C.; Huang, C.; Lv, C.; Zheng, C.; Liu, D.; Zhou, F.; Huang, F.; Hu, F.; Ge, H.; Wei, H.; Lin, H.; Tang, J.; Yang, J.; Tu, J.; Zhang, J.; Yang, J.; Yang, J.; Zhou, J.; Zhou, J.; Lin, J.; Dang, K.; Bao, K.; Yang, K.; Yu, L.; Deng, L.; Li, M.; Xue, M.; Li, M.; Zhang, P.; Wang, P.; Zhu, Q.; Men, R.; Gao, R.; Liu, S.; Luo, S.; Li, T.; Tang, T.; Yin, W.; Ren, X.; Wang, X.; Zhang, X.; Ren, X.; Fan, Y.; Su, Y.; Zhang, Y.; Zhang, Y.; Wan, Y.; Liu, Y.; Wang, Z.; Cui, Z.; Zhang, Z.; Zhou, Z.; and Qiu, Z. 2025.
\newblock Qwen3 Technical Report.
\newblock arXiv:2505.09388.

\bibitem[{Zhang(2023)}]{dfnn}
Zhang, B. 2023.
\newblock Root Cause Analysis of Communication Network Based on Deep Fuzzy Neural Network.
\newblock \emph{IEEE Access}, PP: 1--1.

\bibitem[{Zhang et~al.(2021)Zhang, Kalander, Zhou, Zhang, and Ye}]{zhang2021influencebasedapproachrootcause}
Zhang, K.; Kalander, M.; Zhou, M.; Zhang, X.; and Ye, J. 2021.
\newblock An Influence-based Approach for Root Cause Alarm Discovery in Telecom Networks.
\newblock arXiv:2105.03092.

\end{thebibliography}
